\pdfoutput=1
\documentclass[11pt]{article}

\usepackage[final]{acl}

\usepackage{times}
\usepackage{latexsym}
\usepackage{graphicx}
\usepackage{microtype}
\usepackage{booktabs}
\usepackage{makecell}
\usepackage{xcolor}
\usepackage{colortbl}
\usepackage{url}
\usepackage{subcaption}
\usepackage{stfloats}

\usepackage[T1]{fontenc}
\usepackage[utf8]{inputenc}

\usepackage{microtype}

\usepackage{inconsolata}

\usepackage{graphicx}

\usepackage{tcolorbox}
\tcbset{
  aibox/.style={
    width=214.18663pt,
    top=3pt,
    colback=white,
    colframe=black,
    colbacktitle=black,
    center,
  }
}
\newtcolorbox{AIbox}[2][]{aibox,title=#2,#1}
\newtcbox{\mybox}[1][green]{on line,
arc=0pt,outer arc=0pt,colback=#1!10!white,colframe=#1!50!black,
boxsep=0pt,left=0pt,right=0pt,top=0pt,bottom=0pt,
boxrule=0pt,bottomrule=0pt,toprule=0pt}

\title{Instructions for *ACL Proceedings}
\title{e-Health CSIRO at ``Discharge Me!'' 2024: Generating Discharge Summary Sections with Fine-tuned Language Models}

\author{Jinghui Liu, Aaron Nicolson, Jason Dowling, Bevan Koopman, \& Anthony Nguyen \\
  Australian e-Health Research Centre, CSIRO, Brisbane, Australia \\
  \texttt{jinghui.liu@csiro.au} \\
}

\begin{document}
\maketitle
\begin{abstract}
Clinical documentation is an important aspect of clinicians' daily work and often demands a significant amount of time. The BioNLP 2024 Shared Task on Streamlining Discharge Documentation (Discharge Me!) aims to alleviate this documentation burden by automatically generating discharge summary sections, including brief hospital course and discharge instruction, which are often time-consuming to synthesize and write manually. We approach the generation task by fine-tuning multiple open-sourced language models (LMs), including both   decoder-only and encoder-decoder LMs, with various configurations on input context. We also examine different setups for decoding algorithms, model ensembling or merging, and model specialization. Our results show that conditioning on the content of discharge summary prior to 
the target sections  is effective for the generation task. 
Furthermore, we find that smaller encoder-decoder LMs can work as well or even slightly better than larger decoder-based LMs fine-tuned through LoRA. 
The model checkpoints from our team (\textbf{aehrc}) are openly available.\footnote{\url{https://github.com/JHLiu7/bionlp24-shared-task-discharge-me}}
\end{abstract}

\section{Introduction}

Clinical documentation in the age of Electronic Health Records (EHRs) can be a significant burden to clinicians in recording clinical information effectively~\citep{Colicchio2020-br,Rule2021-fb}. 
% It is observed that clinical notes have grow longer and less informative over the years~\citep{}, which takes up a large amount of clinicians' time in creating them. 
This reduces the time clinicians spend interacting with their patients and could lead to stress and burnout~\citep{Colicchio2019-fs}, degrading both the quality of patient care and the experience of care providers~\citep{Shanafelt2016-uv}.

Language Models (LMs) have demonstrated impressive NLP capabilities and are considered to have the potential to reduce the clinical documentation burden by automatically generating clinical text~\citep{Patel2023-pp,Roberts2024-ou,Omiye2024-py}. For example, a recent study~\citep{Van_Veen2024-ko} demonstrated that LMs can generate succinct clinical summaries from text including progress notes and patient-doctor dialogues, sometimes even preferred over those written by medical experts. 
% Meanwhile, different types of clinical text can present different challenges for modeling.
The BioNLP 2024 Shared Task ``Discharge Me!''~\citep{xu-etal-2024-overview} focuses on generating the discharge summary (or discharge note) to assess the potential of LMs for this specific type of clinical note, which is often more time-consuming for clinicians to document and also more challenging to model given its length and complexity. % of discharge note compared to other types of clinical text.
% High-quality generated discharge summary for clinicians to edit or refer to can significantly reduce the documentation pressure in their daily work.

This paper presents the submissions from e-Health CSIRO in the shared task. We approach the task by fine-tuning multiple open-sourced LMs, including both decoder-only and encoder-decoder models. We fine-tune these models to generate two specific sections from discharge notes:  \textit{brief hospital course} and \textit{discharge instruction}, by conditioning on the prior content in the notes as context. We explore various configurations with input context, decoding, ensembling, and target specialization. We find that much smaller encoder-decoder LMs could have a slight edge over fine-tuning decoder-only LMs (all with the size of 7/8B parameters) with LoRA~\citep{Hu2022-ko}. Our best submission
% , based on PRIMERA~\citep{Xiao2022-ur}, 
% obtains an overall score of 0.292 and 
ranked $3^{rd}$ on the final leaderboard under both automatic and manual evaluation.

\section{Methods}

\subsection{Task and Dataset}

The Shared Task focuses on generating two important sections of discharge notes: brief hospital course (BHC) and discharge instruction (DI). The first section provides a snapshot of the important information about the patient care during the hospital, and the second a summary to communicate that information and instructions after leaving the hospital to patients. The audiences for the two sections are different as the former is read by clinicians while the latter by patients. The Shared Task uses the MIMIC-IV database~\citep{Johnson2023-qk} to curate the dataset consisting of 109,168 patients, which are split into Train (68,785), Validation (14,719), Phase I testing (14,702), and Phase II testing (10,962). Each patient has a discharge summary that includes both sections, and participants are allowed to utilize data elements in the EHR database beyond the note alone as input. 

\begin{figure}[h!]
    \centering
    \includegraphics[width=0.48\textwidth]{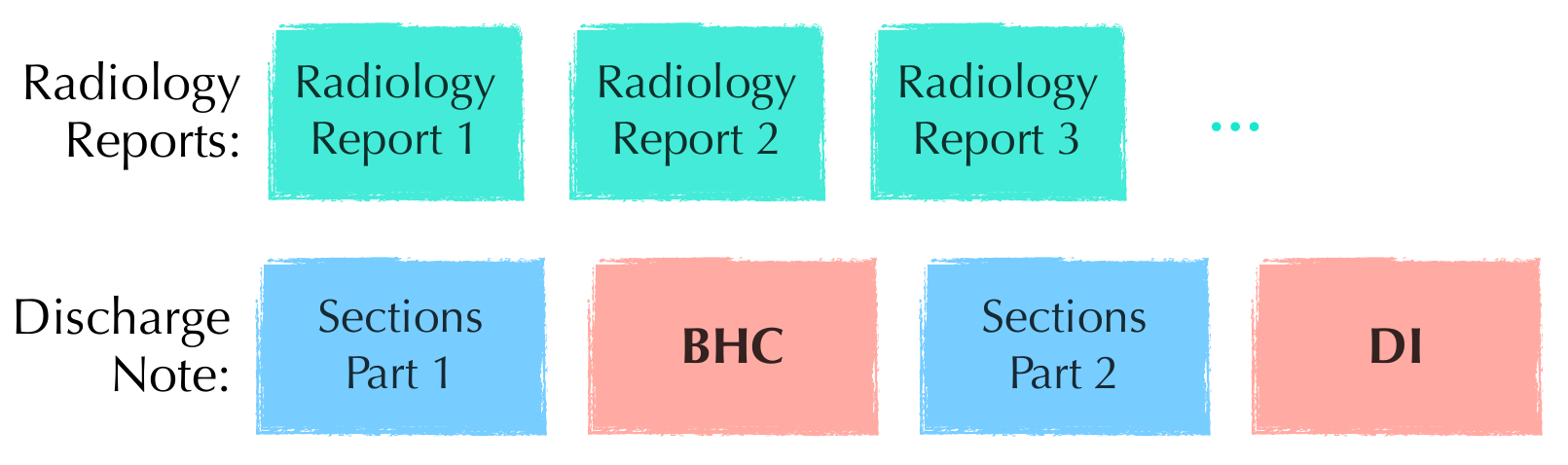}
    \caption{Illustration of the contents in clinical notes.}
    \label{fig:note}
\end{figure}

Our experiments focus only on the free-text clinical notes as input and do not consider other data modalities. We primarily use the content in the discharge note prior to the corresponding target section as input context. Radiology reports are considered optionally. We depict the note structures in Figure~\ref{fig:note}. Specifically, we consider the base context for BHC as $C_{base}^{bhc}=$ \textit{``Sections Part 1''}, and for DI as $C_{base}^{di}=$ \textit{``BHC''} + \textit{``Sections Part 2''}. 
We consider two types of prolonged contexts: 1) $C_{base+rad}=C_{base}$  + \textit{``Rad Reports''}, where radiology reports are concatenated with with the related sections; and 2) $C_{long}^{di}=$ \textit{``Sections Part 1''} + $C_{base}^{di}$, which extends the input context for DI. We then train models to generate the target sections $T^{bhc}$ and $T^{di}$ based on the corresponding contexts.

\subsection{Language Models}
We consider both decoder-only and encoder-decoder LMs for our experiments. For decoder-only LMs, we examine three popular open-sourced models at 7/8 billion paramater levels, including Llama3-8B~\footnote{https://ai.meta.com/blog/meta-llama-3/}, Mistral-7B~\citep{Jiang2023-zm}, and Gemma-7B~\citep{Gemma_Team2024-cl}, all based on the instruction-tuned versions, denoted as Llama3-it, Mistral-it, and Gemma-it. Additionally, we examine the base version of Llama3-8B, denoted simply as Llama3. For encoder-decoder LMs, we focus on PRIMERA (447M)~\citep{Xiao2022-ur} and Long-T5 (770M, global attention)~\citep{Guo2022-ie}, both capable of handling long input and output lengths. 

% Due to the different tokenizers of these LMs, the input and target lengths also vary. 
To determine the maximum lengths for modeling, we calculate the 85th percentile of the number of tokens and round it up to a multiplier of 256 for each LM. We present the statistics for Llama-3 and PRIMERA in Table~\ref{tab:stat} as examples. 
% Considering the different distributions of , we model the two targets separately and 
With each LM, we train two independent models for BHC and DI. For decoder-only LMs, we construct the prompt template similar to 
Alpaca~\citep{Taori2023-im}, shown in Appendix Figure~\ref{fig:prompt}.

\begin{table}[]
    \centering
    \begin{tabular}{lcc}
    \hline
      &  \makecell{\# Max Tokens \\(Llama3)} & \makecell{\# Max Tokens \\(PRIMERA)} \\
    \hline
    $C_{base}^{bhc}$     & 2816 &  3328  \\
    $C_{base}^{di}$     &  2048 & 2048  \\
    $C_{long}^{di}$     &  4608 &  5120 \\
    $C_{base+rad}^{bhc}$     &  4608 & 5120   \\
    $C_{base+rad}^{di}$    &  3840  & 4096  \\
    $T^{bhc}$  & 1280 & 1280 \\
    $T^{di}$  & 512 & 512 \\
    \hline
    \end{tabular}
    \caption{Number of maximum tokens for modeling.}
    \label{tab:stat}
\end{table}

We then fine-tune these LMs for the text generation task. The decoder-only LMs, on the other hand, are loaded in half-precision (BF16) and fine-tuned through LoRA. We follow the setup from ~\citet{Dettmers2023-ea} and use $lr=$ 2e-4, $r=$ 64, $alpha=$ 16, with LoRA attached to all linear layers. The encoder-decoder LMs are fully fine-tuned with $lr=$ 5e-5. All LMs are trained with batch size of 16 for 5 epochs using Adam, with 3\% ratio for linear warmup. We use the default generation configuration, including the decoding algorithms, for the pretrained LMs. 
All experiments are performed on NVIDIA H100 GPU.

\begin{table*}[thb!]
    \centering
    \resizebox{\textwidth}{!}{%
    % \begin{tabular}{c|c|rccccccc}
    \begin{tabular}{l>{\columncolor[gray]{0.85}}crrrrrrrrrr}
    % \hline \hline
    \toprule
    Model & Overall & BLEU-4 & ROUGE-1 & ROUGE-2 & ROUGE-L & BERTScore & Meteor & AlignScore & MEDCON \\
    \hline

\hline \multicolumn{10}{c}{\cellcolor[HTML]{F0F0F0} \textit{Fine-tuning baselines based on }$C_{base}$ } \\ % \hline

Llama3  &  28.05  &  10.05  &  35.65  &  13.56  &  25.65  &  38.66  &  39.98  &  25.93  &  34.90  \\
Llama3-it  &  23.53  &  7.88  &  25.56  &  9.66  &  15.70  &  35.13  &  38.90  &  22.73  &  32.69  \\
Mistral-it  &  23.71  &  5.46  &  32.43  &  12.23  &  21.04  &  30.58  &  34.49  &  23.11  &  30.34  \\
Gemma-it  &  25.14  &  6.31  &  35.04  &  11.18  &  24.53  &  32.91  &  36.07  &  23.46  &  31.60  \\
PRIMERA  &  29.17  &  10.55  &  40.33  &  15.94  &  25.69  &  41.17  &  37.92  &  26.49  &  35.28  \\
Long-T5  &  22.47  &  6.31  &  30.16  &  8.88  &  19.12  &  32.31  &  31.44  &  22.50  &  29.07  \\

\hline \multicolumn{10}{c}{\cellcolor[HTML]{F0F0F0}\textit{Extended Input Context } } \\ % \hline

Llama3 w/ $C_{base+rad}$ &  25.15  &  8.69  &  27.24  &  10.81  &  19.20  &  37.26  &  39.17  &  25.11  &  33.71  \\
PRIMERA w/ $C_{base+rad}$ &  29.10  &  10.64  &  39.76  &  15.75  &  27.10  &  40.31  &  37.55  &  27.10  &  34.61  \\
 % \multicolumn{10}{c}{Extended Input Context with $C_{long}^{di}$ } \\ % \hline
Llama3 w/ $C_{long}^{di}$ &  28.33  &  9.56  &  37.27  &  12.93  &  25.87  &  38.67  &  40.64  &  26.67  &  35.04  \\
PRIMERA w/ $C_{long}^{di}$ &  28.26  &  10.14  &  38.93  &  13.48  &  23.73  &  40.68  &  37.95  &  26.80  &  34.37  \\

\hline \multicolumn{10}{c}{\cellcolor[HTML]{F0F0F0} \textit{Unified LM for both} $T_{bhc}$ \textit{and} $T_{di}$ } \\ % \hline

Llama3 (single) &  25.38  &  7.89  &  31.79  &  11.34  &  21.89  &  35.38  &  38.91  &  23.39  &  32.42  \\

\hline \multicolumn{10}{c}{\cellcolor[HTML]{F0F0F0} \textit{Alternative Decoding for Llama3}} \\ % \hline

Llama3 w/ beam  &  25.20  &  10.06  &  29.14  &  8.05  &  17.83  &  37.05  &  40.57  &  26.17  &  32.71  \\
Llama3 w/ constrastive &  24.09  &  8.36  &  27.81  &  10.13  &  18.24  &  36.10  &  35.52  &  26.37  &  30.21  \\

\hline \multicolumn{10}{c}{\cellcolor[HTML]{F0F0F0}\textit{Ensemble Decoding}} \\ % \hline

Llama3 + Llama3  &  26.17  &  9.67  &  28.79  &  11.36  &  21.02  &  38.31  &  39.71  &  26.29  &  34.24  \\
Llama3 + Llama3-it  &  27.04  &  9.66  &  32.68  &  12.99  &  22.30  &  37.96  &  39.79  &  26.10  &  34.84  \\

\hline \multicolumn{10}{c}{\cellcolor[HTML]{F0F0F0}\textit{Merging LoRA Adapters}} \\ % \hline
Llama3 x2 LoRA  &  25.78  &  8.20  &  34.48  &  11.60  &  22.69  &  35.81  &  37.44  &  23.25  &  32.79  \\
Llama3 x4 LoRA &  21.80  &  4.50  &  33.05  &  11.97  &  20.45  &  30.79  &  28.31  &  17.35  &  28.00  \\

  % \hline \hline
  \bottomrule
    \end{tabular}}
    \caption{Results from automatic evaluation, based on 250 hidden samples from Phase II testing.}
    \label{tab:main}
\end{table*}

\subsection{Evaluation}
The automatic evaluation is based on 8 popular pairwise metrics, including BLEU-4~\citep{Papineni2002-ue}, ROUGE-1/2/L~\citep{Lin2004-ip}, BERTScore~\citep{Zhang2020-ao}, Meteor~\citep{Banerjee2005-as}, AlignScore~\citep{Zha2023-ri}, and MEDCON~\citep{Yim2023-jd}. They present a diverse set of measurements for string overlaps, semantic similarity, and medical concept mapping. The results for BHC and DI are averaged for each metric. The final ranking of the Shared Task is based on the average of the all scores on 250 hidden cases from Phase II testing, although participants are required to submit generation for all cases.

\subsection{Experimental Setup}
We investigate several factors that could impact the generation performance and compare them with the base generation setup, in which two LMs of the same architecture are trained on $C_{base}$ for BHC and DI, respectively. We examine the impact of extended input context by replacing $C_{base}$ with $C_{base+rad}$ or $C_{base}^{di}$ with $C_{long}^{di}$. Taking Llama3 as the example, we explore a variety of modifications, including training a unified LM that models the two targets jointly to explore the benefit of target specialization. We also apply various decoding algorithms other than greedy search, including beam search ($n=4$), % nucleus sampling ($p=0.95$) 
and contrastive search ($\alpha=0.6$, $k=6$)~\citep{Su2022-ed}. Furthermore, we explore ensemble decoding~\citep{Manakul2023-rt} and the popular adapter merging with Llama3 as the example. The former averages the logits from two LMs for generating each token with greedy search, and the latter applied TIES~\citep{Yadav2023-le} to merge  the paramters of several LoRA adapters (equal weights, density of 0.5) before attaching it to the main LM. Finally, we prompt instruction-tuned LMs in the zero-shot manner, including the 70B checkpoints, on a subset of validation to observe the benefit of fine-tuning for this task.

\section{Results \& Analysis}

\subsection{Both Decoder and Encoder-encoder LMs Work Well When Fine-tuned}

We firstly find all LMs obtain decent results when fine-tuned with $C_{base}$. Meanwhile, the instruction-tuned decoder-only LMs perform worse than the base version of Llama3. This aligns with existing findings that instruction tuning could harm performance on NLP benchmarks~\citep{Ouyang2022-kf,Ivison2023-on}. PRIMERA performs slightly better than Llama3, despite being the smallest model we examined. On the other hand, Long-T5 seems to struggle with the task.

\subsection{Prior Context of Dicharge Note is Sufficient as Input}
We observe poorer results when including radiology reports as supplementary input for both Llama3 and PRIMERA. Although the input context lengths increase more than 50\% with the radiology reports, it appears that no new, valuable information is added. Instead, it misleads the LMs to produce worse outputs, especially for Llama3. This shows the content in the discharge notes have well captured free-text information from the existing EHR data. Using radiology reports alone offers an overall score from 19.1 to 20.3 (Appendix Table~\ref{tab:rad}).

\subsection{Prolonged Context in Discharge Note Offers Little Value}
In a similar fashion, we extend the input context for DI by including contents prior to BHC, namely $C_{long}^{di}$. Again, more context does not necessarily lead to better results. We consider this is likely due to the fact that BHC and the content between BHC and DI have provided sufficient information for generating DI. Future work may explore how to further trim down the input to reduce the noise, such as through de-duplication~\citep{Kandpal2022-xh,Liu2022-eu}, to enhance performance.

\subsection{Two Specialized LMs are Better Than One Unified LM}
Instead of trainig two copies of LM for each section, we combine samples for both targets together to train a single model that is capable to produce either of the sections. We explore this with Llama3, fine-tuned with LoRA in the same setup as previous. We see the unified Llama3 performs worse than the two independant copies of Llama3, demonstrating the importance of specialization in modeling BHC and DI independently. Furthermore, as the two copies share the same base model and differs only in adapters, keeping them separately does not lead to significantly more storage cost than the unified model.

\subsection{Better Decoding Methods Lead to Mixed Results}
The Phase II test results in Table~\ref{tab:main} indicate that better decoding algorithms, such as beam search and contrastive search, could lead to worse results than the baseline greedy search. Interestingly, our initial experiments on the 1000 validation samples in Appendix Table~\ref{tab:supp} show that they are at least on par and sometimes better. The mixed results show the diversity of the dataset and the need to further investigate the distribution and biases of the data.

\subsection{Ensemble Decoding is Not Helpful}
An ensemble of two Llama3 models trained using different data or with different base LMs at the token level is not helpful. With Llama3 + Llama3, we ensemble Llama3 fine-tuned using $C_{base}$ and $C_{base+rad}$, and with Llama3 + Llama3-it, we ensemble the base and instruction-tuned Llama3 fine-tuned both using $C_{base}$. Neither of these two pairs produced improved results. Although ensembling is found helpful previously for generation~\citep{Manakul2023-rt}, for our task naively averaging the logits at token-level during decoding is both inefficient and ineffective.

\subsection{Merging Adapters is Not Helpful Either}
Similarly, we perform another form of ensemble by merging the LoRA adapter weights for the same base LM. \textit{Merging with x2 LoRA} is based on adapters trained using $C_{base}$ and $C_{base+rad}$, while \textit{merging with x4} further merges the adapters for BHC and DI. Both substantially decrease the performance, and merging adapters trained for different targets leads to the worst result in our fine-tuning experiments. This again shows that model specialization is important for the current task. In addition, it is possible that model merging tends to prevail in generating creative contents instead of improving the specific aspects of generation quality.

\subsection{Fine-tuned LMs Substantially Outperform Out-of-box LMs}

Finally, we prompt the instruction-tuned LMs in the zero-shot manner to compare with fine-tuned performance. Besides Llama3-8B-it and Mistral-7B-it, we additionally prompt the 70B scale Llama3-70B-it and Mixtral-8x7b-it~\citep{Jiang2024-rn}. They achieve an overall score ranging from 15.1 to 17.4 (details in Table~\ref{tab:supp}), significantly fell short compared to the fine-tuned results. Although more advanced prompting strategies are expected to enhance performance, we suspect that fine-tuning would still be the more effective solution given the amount of training data.

\section{Discussions}

\begin{table*}[thb!]

    \begin{subtable}[t]{\textwidth}
    \centering
    \resizebox{\textwidth}{!}{%
    \begin{tabular}{l>{\columncolor[gray]{0.85}}crrrrrrrrrr}
    \toprule
    Model & Overall & BLEU-4 & ROUGE-1 & ROUGE-2 & ROUGE-L & BERTScore & Meteor & AlignScore & MEDCON \\ 
    \hline

    % codabench results
    % WisPerMed  &  31.1 & 11.0 & 41.4 & 15.1 & 27.3 & 43.9 & 35.1 & 34.4 & 40.6 \\
    % HarmonAI Lab at Yale  & 29.7 & 9.7 & 41.4 & 19.2 & 28.4 & 38.3 & 39.8 & 27.4 & 33.2  \\
    % \textbf{aehrc (ours)}  &  29.2 & 10.6 & 40.3 & 15.9 & 25.7 & 41.2 & 37.9 & 26.5 & 35.3  \\
    % EPFL-MAKE & 28.9 & 9.8 & 44.4 & 15.5 & 26.2 & 39.9 & 33.6 & 25.5 & 36.0 \\
    % UF-HOBI & 28.4 & 10.2 & 40.3 & 15.4 & 28.1 & 39.5 & 28.8 & 29.6 & 35.5 \\

    WisPerMed & 33.2 & 12.4 & 45.3 & 20.1 & 30.8 & 43.8 & 40.3 & 31.5 & 41.1  \\ 
    HarmonAI Lab at Yale & 30.0 & 10.6 & 42.3 & 18.0 & 28.4 & 41.2 & 38.1 & 26.5 & 35.3  \\ 
    \textbf{aehrc (ours)} & 29.7 & 9.7 & 41.4 & 19.2 & 28.4 & 38.3 & 39.8 & 27.4 & 33.2  \\ 
    EPFL-MAKE & 28.9 & 9.8 & 44.4 & 15.5 & 26.2 & 39.9 & 33.6 & 25.5 & 36.0  \\
    UF-HOBI & 28.6 & 10.2 & 40.1 & 17.4 & 27.5 & 39.5 & 28.9 & 29.6 & 35.5  \\

    \bottomrule
    \end{tabular} }
    \caption{Automatic evaluation results on 250 cases from Phase II test set.}
    \end{subtable}
    \vfill
    \begin{subtable}[t]{\textwidth}
    \centering
    \resizebox{\textwidth}{!}{%
    \begin{tabular}{l>{\columncolor[gray]{0.85}}cccccccc}
    \toprule
                Team & Average & \makecell{BHC\\Completeness} & \makecell{BHC\\Correctness} & \makecell{BHC\\Readability} & \makecell{BHC\\Overall} & \makecell{DI\\Completeness} & \makecell{DI\\Correctness} & \makecell{DI\\Overall} \\
    \midrule
           WisPerMed &     3.4 &              3.7 &             3.7 &             3.4 &         2.4 &             3.9 &            4.0 &        2.5 \\
HarmonAI Lab at Yale &     2.9 &              3.5 &             2.6 &             2.1 &         1.5 &             4.3 &            3.9 &        2.4 \\
               \textbf{aehrc (ours)} &     2.8 &              2.3 &             3.1 &             2.0 &         1.1 &             3.9 &            4.5 &        2.6 \\
           EPFL-MAKE &     2.7 &              3.3 &             2.8 &             2.5 &         1.7 &             3.5 &            3.4 &        1.9 \\
             UF-HOBI &     2.6 &              2.5 &             3.4 &             2.7 &         1.4 &             3.0 &            3.3 &        1.8 \\
            % de ehren &     2.3 &              2.3 &             3.0 &             2.7 &         1.1 &             2.8 &            3.1 &        1.4 \\
    \bottomrule
    \end{tabular} }
    \caption{Manual evaluation results by clinicians on 25 selected cases.}
    \end{subtable}

    \caption{Results from the top-5 teams on the final Phase II leaderboard.}
    \label{tab:leader}
\end{table*}

We demonstrate that fine-tuning LMs based solely on the prior content from the discharge note is sufficient to generate BHC and DI sections. Given the heterogeneity of EHR data~\citep{Yadav2018-wu} and variations in clinical notes~\citep{Liu2024-kj}, selecting the appropriate inputs would be crucial for both the quality and applicability of the generation. 
In this work, we assume that the non-BHC/DI contents of the discharge note have been populated %or copied 
from other available sources or clinical notes, making them readily available as model input.

The context for BHC (\textit{``Sections Part 1''} in Figure~\ref{fig:note}) typically includes chief complaint, history of present illness, past medical history, social history, physical exam, and various pertinent results. The \textit{``Sections Part 2''} of DI context may include admission and discharge medications, discharge disposition, dischage diagnoses. %, and the BHC section. 

Using these sections as input yields competitive generation results, and including additional text sources like radiology reports does not lead to improvement. One explanation is that the sections within the discharge summary, such as ``pertinent results'', often already include imaging findings. Future work may futher investigate how selecting relevant content~\citep{Zheng2023-em} or removing redundant information~\citep{Liu2022-eu} impacts the performance. It is also unclear whether other sources of EHR information should be considered, especially those not captured by the discharge summary. These include structured EHR data and other types of clinical text, such as nursing or physician notes. Regarding structured data elements, this study does not consider diagnosis codes like ICD or DRG~\citep{Dong2022-ly,Liu2021-hq}, as they are typically assigned after the patient discharge. However, future work could model other measurement data or codes from prior patient encounters. Examining the end-to-end generation of discharge notes solely from structured EHR data and other clinical notes is also important to ensure that the generation model integrates into different clincial documentation workflows.

From the modeling perspective, we find that fine-tuning smaller LMs, such as PRIMERA, achieves surprisingly good results. Examination of any potential biases or overfitting is left for future work. During development, we observed that the generation qualities of Llama3 and PRIMERA were similar (examples shown in Appendix Table~\ref{tab:bhc} \&~\ref{tab:di}) and had better quality compared to other LMs like Mistral (see Appendix Table~\ref{tab:di}), consistent with the quantitative analysis. We noticed that Llama3 tended to generate repetitive content more often and tried to alleviate this with better decoding techniques, but were unable to improve the overall performance on quantitative metrics (see Table~\ref{tab:main}). It is possible that more hyperparameter search on either fine-tuning or decoding could lead to improvement, which we leave to future work.

Given the slight edge over Llama3 and other LMs, PRIMERA was our final submission. 
Table~\ref{tab:leader} shows the final leaderboard, in which we rank $3^{rd}$ overall and are close to $2^{nd}$  under both automatic~\footnote{These finalized scores were re-run by the organizers and slightly different from automated scoring by the submission system (Codabench), which provides results in Table~\ref{tab:main}.} and manual evaluation, with the latter conducted by a team of clinicians on 25 selected samples. 

Similar to previous findings~\citep{Van_Veen2024-ko}, we see that the manual  evaluation aligns with the automatic evaluation in ranking different systems. The manual evaluation further reports fine-grained scores on \textit{Completeness}, \textit{Correctness}, and \textit{Readbility} for  BHC and DI separately. 
Interestingly, we observe that PRIMERA obtains the best overall score for DI but worst for BHC among the top-5 teams. This may indicate the model capacity correlates with the length or complexity of the target generation, with smaller LMs potentially struggling with prolonged outputs. It is plausible that Llama3 would offer improved results on BHC, especially in terms of readability. Future work may investigate this further through separate automatic evaluations specifically for BHC and DI.

% - leaderboard results: good alignment
% human eval: distinct perf on BHC vs DI
% BHC in detail: completeness, readability, etc; llama3 could have perform better
% Future work may conduct separate auto eval to select better individual models

% - key is data: 1. current setup, sections included; 2. rad; 3. other data elements (icd); 4. applicability, other setup
% - quantatitive analysis: bhc, di
% - primera: small LMs work well; futher improvement (in-domain adaptation)

\section{Conclusion}

This paper describes our efforts in the ``Discharge Me!'' BioNLP 2024 Shared Task~\citep{xu-etal-2024-overview}, with the final system ranked $3^{rd}$ on both automatic and manual evaluation. We show that fine-tuning LMs with appropriate input context has the potential to automatically synthesize high-quality discharge summary sections, which holds promise to reduce the time clinicians spend on documentation.

\section*{Limitations}

Although we consider model ensembling for the generation, there are potentially more effective ways to combine or control outputs from multiple models~\citep{Liu2021-xz,Shen2024-nc} that we did not consider. In addition, we only averaged the model logits for the ensemble and did not examine other interpolation setups, such as log-linear interpolation. Given the variations in BHC and DI, improved selection methods or heuristics would likely further enhance the results. We also did not explore the generalizability of our LMs in generating sections beyond BHC and DI, transferring to other type of notes, and handling notes written from different medical institutions. Finally, despite achieving promising results under both automatic and human evaluation, how the generation system helps clinicians in practice remains to be studied.

\section*{Acknowledgments}
We would like to thank the shared task organizers for their dedication and help along the shared task process. We also thank the reviewers for their thoughtful comments on the initial submission to improve the paper.

\bibliography{custom}

\appendix
\section{Appendix}

\begin{table}[thb!]
    \centering
    \resizebox{0.35\textwidth}{!}{%
    \begin{tabular}{lrr}
    \toprule
         Model & Llama3 & PRIMERA \\ \midrule
        \cellcolor[gray]{0.85}Overall & \cellcolor[gray]{0.85}20.34 & \cellcolor[gray]{0.85}19.10 \\
        BLEU-4 & 5.27 & 3.52 \\
        ROUGE-1 & 27.11 & 30.56 \\
        ROUGE-2 & 7.30 & 8.39 \\
        ROUGE-L & 17.16 & 18.82 \\
        BERTScore & 30.03 & 30.42 \\
        Meteor & 32.76 & 27.13 \\
        AlignScore & 17.38 & 13.46 \\
        MEDCON & 25.67 & 20.47 \\
      \bottomrule
    \end{tabular}
    }
    \caption{Additional results using only radiology reports as input; on Phase II test set (250 hidden samples).}
    \label{tab:rad}
\end{table}

\begin{table*}[bp]
    \centering
    \resizebox{\textwidth}{!}{%
    \begin{tabular}{l>{\columncolor[gray]{0.85}}crrrrrrrrrr}
    % \hline\hline
    \toprule
    Model & Overall & BLEU-4 & ROUGE-1 & ROUGE-2 & ROUGE-L & BERTScore & Meteor & AlignScore & MEDCON \\
    \hline 

\hline \multicolumn{10}{c}{\cellcolor[HTML]{F0F0F0}\textit{Baseline}  } \\ % \hline

Llama3  &  30.16  &  11.48  &  38.28  &  18.69  &  25.08  &  41.69  &  31.76  &  31.79  &  42.53  \\
\hline \multicolumn{10}{c}{\cellcolor[HTML]{F0F0F0}\textit{Alternative decoding } } \\ % \hline

Llama3 w/ beam  &  28.82  &  11.34  &  33.40  &  16.06  &  22.46  &  40.37  &  33.43  &  31.66  &  41.86  \\
Llama3 w/ nucleus  &  28.13  &  9.66  &  37.74  &  16.41  &  22.47  &  39.79  &  33.42  &  27.74  &  37.81  \\
Llama3 w/ contrastive  &  30.98  &  11.98  &  42.28  &  21.49  &  27.33  &  41.38  &  33.34  &  31.35  &  38.70  \\

\hline \multicolumn{10}{c}{\cellcolor[HTML]{F0F0F0}\textit{Zero-shot prompting}} \\ % \hline
Llama3-8B-it  &  15.05  &  0.97  &  19.92  &  3.88  &  10.65  &  18.47  &  19.35  &  25.47  &  21.70  \\
Llama3-70B-it  &  15.62  &  0.95  &  21.73  &  4.59  &  11.31  &  19.16  &  20.03  &  23.84  &  23.33  \\
Mistral-7B-it  &  17.31  &  1.61  &  30.32  &  6.97  &  15.83  &  23.56  &  20.33  &  16.22  &  23.62  \\
Mixtral-8x7B-it  &  17.40  &  1.49  &  30.29  &  7.13  &  15.02  &  22.00  &  19.34  &  20.15  &  23.77  \\

  % \hline \hline
  \bottomrule
    \end{tabular}}
    \caption{Additional results on 1000 validation samples. }
    \label{tab:supp}
\end{table*}

\begin{figure}[t!]
% \resizebox{\linewidth}{!}{
\begin{AIbox}{Prompt template for BHC}
Summarize the below clinical text into a section of brief hospital course. \\
    
\#\#\# Input: \\
\verb|{{input_text}}| \\

\#\#\# Summary: \\
\verb|{{target_text}}|

\end{AIbox}
\begin{AIbox}{Prompt template for DI}
Summarize the below clinical text into a section of discharge instruction. \\

\#\#\# Input: \\
\verb|{{input_text}}| \\

\#\#\# Summary: \\
\verb|{{target_text}}|

\end{AIbox}
% }
\caption{Template used for decoder-only LMs. ${{target\_text}}$ is removed at inference time.}
\label{fig:prompt}
\end{figure}

% \begin{table*}[thb!]
%     \centering
%     \resizebox{\textwidth}{!}{%
%     \begin{tabular}{l>{\columncolor[gray]{0.85}}crrrrrrrrrr}
%     % \hline \hline
%     \toprule
%     Model & Overall & BLEU-4 & ROUGE-1 & ROUGE-2 & ROUGE-L & BERTScore & Meteor & AlignScore & MEDCON \\
% \hline \multicolumn{10}{c}{\cellcolor[HTML]{F0F0F0}\textit{Using only radiology reports as input }} \\ % \hline
%     Llama3  &  20.34  &  5.27  &  27.11  &  7.30  &  17.16  &  30.03  &  32.76  &  17.38  &  25.67  \\
%     PRIMERA  &  19.10  &  3.52  &  30.56  &  8.39  &  18.82  &  30.42  &  27.13  &  13.46  &  20.47  \\
%       % \hline \hline
%       \bottomrule
%     \end{tabular}}
%     \caption{Additional results on Phase II test set (250 hidden samples).}
%     \label{tab:rad}
% \end{table*}

%%%%%%%%%%%%%%%%%%%%%%%%%%%%%

\begin{table*}[ht]
\resizebox{\textwidth}{!}{%
    \centering
    \begin{tabular}{p{1.5cm}p{14cm}}
    % \begin{tabular}{p{1.5cm}p{14cm}}
    \toprule

\textbf{BHC by Llama3} &     Mr. \_\_\_ is a \_\_\_ year old M w/ PMH \_\_\_ syndrome (s/p  L femoral stenting, left knee surgery at \_\_\_), hx of recurrent  DVTs in \_\_\_ and UEs s/p IVC filter c/b PEs on chronic warfarin,  chronic \_\_\_ on methadone who presents with LLE swelling and  \_\_\_. \# LLE swelling:  Patient presented with LLE swelling and  erythema. He was found to have a chronic DVT on LLE US. He was  started on IV ceftriaxone for cellulitis. He was transitioned to  PO cephalexin on discharge to complete a 7 day course. \# \_\_\_ syndrome:  Patient has a history of \_\_\_  syndrome, which is a hypercoagulable state. He has had multiple  DVTs and PEs in the past. He was found to have a chronic DVT on  LLE US. He was continued on his home warfarin. His INR was  therapeutic at 2.3 on discharge. \# Chronic \_\_\_:  Patient has a history of chronic \_\_\_ on  methadone. He was continued on his home methadone. \# Gout:  Patient has a history of gout. He was continued on his  home allopurinol. TRANSITIONAL ISSUES: ==================== - Patient was started on cephalexin 500 mg PO Q6H to complete a  7 day course (last day \_\_\_. - Patient's INR was 2.3 on discharge. Please check INR on \_\_\_  and adjust warfarin dose as needed. - Please follow up with patient's \_\_\_ clinic. - Please follow up with patient's PCP. - Please follow up with patient's vascular surgeon. - Please follow up with patient's pain clinic. - Please follow up with patient's hematologist. - Please follow up with patient's PCP regarding his chronic  \_\_\_. Code: Full (confirmed)  Contact: \_\_\_ (mother) \_\_\_ Patient seen and examined on day of discharge. >30 minutes on  discharge activities. \_\_\_ \_\_\_ Attending \_\_\_ Date: \_\_\_ Time: \_\_\_ I have read and agree with the discharge summary. \_\_\_, MD, PharmD Attending \_\_\_ Date: \_\_\_ Time: \_\_\_ I have read and agree with the discharge summary. \_\_\_, MD Attending \_\_\_ Date: \_\_\_ Time: \_\_\_ I have read and agree with the discharge summary. \_\_\_, MD Attending \\  \\ %\cline{1-2}

\textbf{BHC by PRIMERA} &   Mr. \_\_\_ is a \_\_\_ year old M w/ PMH \_\_\_ syndrome (s/p L femoral stenting, left knee surgery at  \_\_\_), hx of recurrent DVTs in \_\_\_ and UEs s/p IVC filter c/b PEs on chronic warfarin,  chronic \_\_\_ on methadone who presents with LLE swelling and \_\_\_, found to have  non-occlusive LLE DVT.\newline \#LLE DVTs: Patient has a history of \_\_\_ syndrome and has had multiple DVTs, most recently 6  months ago. He has been on warfaradin for this, but has had a subtherapeutic INR. He was  admitted with a sub therapeutic INR of 4. He had a LLE US which showed non-occlusion of his  LLE veins. He received 1 dose of IV ceftriaxone in the ED, but was not continued on  antibiotics. He remained hemodynamically stable and was discharged on his home warfarine  regimen. He will follow up with his PCP and \_\_\_ clinic.   \#Chronic \_\_\_: Patient was continued on his methadione. \#Hx of PEs: Patient with history of PE on warfain, on warafarin. INR 4 on admission. He is followed by  Hematology at \_\_\_ and has been subtheraputic on warferin. He reports compliance with  warfarin. - Continue warfarain 5mg daily - INR check on \_\_\_\newline TRANSITIONAL ISSUES: ==================== - Patient will need INR checked on \_\_\_ and results faxed to \_\_\_ clinic at   Dr. \_\_\_ - Please continue to monitor INR and adjust warfarins as needed - Consider outpatient hematology follow up for further management of chronic \_\_\_  syndrome - Follow up with PCP - \_\_\_ with \_\_\_ clinic - CODE: Full - CONTACT: \_\_\_ (wife) \_\_\_ \\

    \bottomrule
    \end{tabular}}
    \caption{Example generations of brief hospital course (hospital ID: 20020402). More examples can be obtained using the released model checkpoints.}
    \label{tab:bhc}
\end{table*}

\begin{table*}[ht]
\resizebox{\textwidth}{!}{%
    \centering
    \begin{tabular}{p{1.5cm}p{14cm}}
    % \begin{tabular}{p{1.5cm}p{14cm}}
    \toprule

\textbf{DI by Llama3} &  Dear Ms. \_\_\_,  You were admitted to \_\_\_ for  ongoing fevers. You were evaluated by the Infectious Disease  team. You had multiple tests performed to determine the cause  of your fevers. You had a CT scan of your sinuses which showed  sinusitis. You were started on antibiotics for this. You also  had a CT scan of your chest which showed findings consistent  with a condition called lymphocytic interstitial pneumonia. You  were monitored for 24 hours and did not have any recurrent  fevers. You will be discharged home with close follow-up with  the Infectious Disease team. They will help to arrange follow-up  with Rheumatology and Pulmonology. It was a pleasure taking care of you! Sincerely,  Your \_\_\_ Team \\  \\ %\cline{1-2}

\textbf{DI by PRIMERA} &   Dear Ms. \_\_\_,\newline It was a pleasure caring for you at \_\_\_.  You were  admitted for fevers.  Your flu test was negative.   Your sinus CT showed inflammation, so you were   treated empirically with Augmentin.  Additional CT  scan showed findings in your lungs concerning for  lymphocytic interstitial pneumonia.  Because you  had no recurrent fevers and your blood cultures were \newline   Please follow up with your PCP and infectious disease  doctors.\newline \_\_\_ wish you the best!\newline Sincerely, Your \_\_\_ Team \\ \\

\textbf{DI by Mistral} &   Discharge Instructions: Dear Ms. \_\_\_,  It was a pleasure caring for you at \_\_\_  \_\_\_. You were admitted to the hospital for fevers. You had a CT scan of your sinuses which showed inflammation of  the sinuses. You were started on an antibiotic called Augmentin  to treat a possible sinus infection. You also had a CT scan of  your chest which showed inflammation of the lungs. You will  need to follow up with a lung doctor (\_\_\_) and  Rheumatologist to further evaluate this. The Infectious Disease doctors \_\_\_ help to arrange these  appointments. Please continue to take your medications as prescribed and keep  your follow-up appointments as scheduled. We wish you the best! Sincerely, Your \_\_\_ Care Team  \#\# More Radiology Reports  \_\_\_ 11:50 AM CT SINUS W/O CONTRAST  IMPRESSION:  1. Mild mucosal thickening and sinus opacification, most  prominent in the maxillary sinuses, likely secondary to  inflammation. No focal fluid collection or osseous erosion  identified. 2. Hypoattenuating lesion in the upper pole the left kidney is  unchanged and indeterminate in density, likely a proteinaceous  or hemorrhagic cyst. Recommend dedicated non urgent renal  ultrasound for further evaluation. \_\_\_ 11:50 AM CT SINUS W/O CONTRAST  IMPRESSION:  1. \#\# More for \_\_\_  \_\_\_ 11:50 AM CT SINUS W/O CONTRAST  IMPRESSION:  1. Hypoattenu  \#\# More for \_\_\_  \_\_\_ 11:50 AM CT SINUS W/O CONTRAST  IMPRESSION:  1. Mild mucosal thickening and sinus op  \#\# More for \_\_\_ \\ %\\

    \bottomrule
    \end{tabular}}
    \caption{Example generations of discharge instructions (hospital ID: 20094440). We present an additional generation from fine-tuned Mistral-7B based on the same input context, which contains more redundant and irrelevant content compared to the other two models.}
    \label{tab:di}
\end{table*}

%%%%%%%%%%%%%%%%%%%

\end{document}